%% file: iros2016.tex
\documentclass[letterpaper, 10 pt, conference]{ieeeconf}

\IEEEoverridecommandlockouts
\usepackage{siunitx}
\usepackage{booktabs}
\usepackage{amsmath}
\usepackage{graphicx}
\usepackage{nameref}
\usepackage{subfig}
\usepackage{placeins}
\usepackage[font=footnotesize,labelfont=bf]{caption}
\usepackage{booktabs}
\usepackage{array}

\usepackage{color,soul}
\usepackage{fixme}
\fxsetup{%
    status=draft,
    theme=color
}

\usepackage[noconfig]{refstyle}
\input{refconfig.cfg}

\usepackage{siunitx}

\usepackage[ruled,linesnumbered]{algorithm2e}

\usepackage{tikz}
\usepackage{pgfplots}
\definecolor{col1}{rgb}{1.00, 0.14, 0.14}
\definecolor{col2}{rgb}{0.13, 0.17, 0.53}
\definecolor{col3}{rgb}{0.99, 0.68, 0.38}
\definecolor{col4}{rgb}{0.67, 0.87, 0.64}

\makeatletter
\let\NAT@parse\undefined
\makeatother
\usepackage[numbers,square]{natbib}
\usepackage{algorithm2e}
\SetAlFnt{\small}
\usepackage{algorithmic}
\algsetup{linenosize=\small}
\usepackage{hyperref}
\usepackage{authorindex}

\title{Urban Scene Segmentation with Laser-Constrained CRFs}
\author{Charika De Alvis \hspace{20mm}    Lionel Ott   \hspace{20mm}  Fabio Ramos%
    \thanks{Charika De Alvis, Lionel Ott and Fabio Ramos are with the
    School of Information Technologies, The University of Sydney,
    Australia.}
}

\begin{document}
\let\chapter\section
\maketitle
\thispagestyle{empty}
\pagestyle{empty}

\begin{abstract}

    Robots typically possess sensors of different modalities,
    such as colour cameras, inertial measurement units, and 3D laser
    scanners. Often, solving a particular problem becomes easier
    when more than one modality is used. However, while there are
    undeniable benefits to combine sensors of different modalities the process tends to be
    complicated. Segmenting scenes observed by the
    robot into a discrete set of classes is a central requirement for
    autonomy as understanding the scene is the first step
    to reason about future situations. Scene segmentation is commonly
    performed using either image data or 3D point cloud data. In
    computer vision many successful methods for scene segmentation are
    based on conditional random fields (CRF) where the maximum a
    posteriori (MAP) solution to the segmentation can be obtained by
    inference. In this paper we devise a new CRF inference method
    for scene segmentation that incorporates global constraints, enforcing the sets of nodes are
    assigned the same class label. To do this efficiently, the CRF
    is formulated as a relaxed quadratic program whose MAP
    solution is found using a gradient-based optimisation approach. The
    proposed method is evaluated on images and 3D point cloud data
    gathered in urban environments where image data
    provides the appearance features needed by the CRF, while the 3D point cloud
    data provides  global spatial constraints over sets of nodes. Comparisons
    with belief propagation, conventional quadratic programming
    relaxation, and higher order potential CRF show the benefits of the
    proposed method.

\end{abstract}

\section{Introduction}
\label{sec:lay}

Scene segmentation is a core competency for many robotic tasks. It
provides the foundation which allows a robot to understand and reason
about its environment. For navigation in urban environments such
information is critical for safety, as it allows the robot to predict
which areas pose a risk due to the presence of dynamic objects.
Typically robots
carry many different sensors, such as cameras, laser scanners,
\mbox{RGB-D}
cameras, etc, which typically observe the environment from slightly
different angles. This variation in view point and modality makes
the optimal combination of sensors very challenging. In this paper we propose a
model which effectively combines multiple modalities. The method is
applied to image segmentation using camera and laser scan data
but is general in nature and applicable to a wide
variety of sensor combinations.

Our method is based on a relaxed quadratic program formulation of CRFs
for scene segmentation which enforces a set of global constraints. Image
data is used to build the CRF graph and potential functions while the
depth data is used to formulate global constraints over sets of nodes in
the CRF. These constraint sets contain all nodes belonging to the same
object, as determined by the depth data and ensure they take the same
label during the inference process. The method finds the MAP
solution using an efficient gradient based algorithm, based on
\cite{Zhang2012}. The main contributions of the paper are:
\begin{itemize}
  \item Novel CRF formulation using global constraints capable of
      enforcing label consistency;
  \item Experimental evaluation of the proposed method for scene
      segmentation using image and 3D laser data gathered by a robotic
      platform.
\end{itemize}

The remainder of the paper is structured as follows. In
\secref{related-work} we give an overview of work related to ours,
before we introduce our method in \secref{method}. In
\secref{experiments} we provide experimental evaluation of our method
before concluding in \secref{conclusion}.

\section{Related Work}
\label{sec:related-work}

In computer vision many successful image segmentation methods are based
on graph cuts \citep{boykov2001} and refinements such as normalised cuts
\citep{shi2000, boykov2006}. Graph cuts represent the image as a graph
and attempt to find the set of edges with minimal cost, that when cut
results in a segmentation of the image. There are other approaches that
work on a similar representation but use a different way of solving the
problem. \citet{felzenszwalb2004} propose a method that uses greedy
local segmentation decisions to obtain accurate global results. A novel
graphical model, associative hierarchical random fields, with
applications to scene segmentation has proposed in \citep{ladicky_2014}.
Stereo vision based scene segmentation is another common method. For
example \citet{he2013} present a method to build a dense 3D semantic
occupancy map of an environment based on semantic labels obtained using
a Markov random field which are used to update the semantic labels of
the map cells. A similar approach is taken in \citep{sengupta2013} using
a CRF for the segmentation task and creating a triangulated mesh of the
environment rather than a voxel grid. All of these approaches use only
image data without any additional outside information.

In robotics there has been a lot of work on scene segmentation using
multiple modalities, such as camera and 3D laser data.
\citet{douillard2007} propose a spatial-temporal CRF method integrating
measurements from a conventional $2D$ laser scanner with images from a
calibrated camera. \citet{munoz2012} extract features from image and
laser data and use these in a classifier to segment the scene. A method
that accumulates image based segmentation results in a $3D$ point cloud
was presented by \citet{hermans2014}. An extension of
\citep{felzenszwalb2004} to RGBD data is presented in \citep{strom2010},
taking advantage of distance and normal information. In
\citep{Aijazi2013} a link-chain clustering method operating on a super
voxel representation of RGB-D data was presented. \citet{xu2014} present
a method using multiple independent classifiers with a sophisticated
fusion framework. A method that exploits both colour and depth
information with the help of a CRF is presented in \citep{zhang_2015}. This
method makes predictions separately on the depth and colour data and
fuses the results using a CRF.


Higher order potentials (HOP) allow encoding additional information
which the unary and pairwise potentials of a CRF cannot represent. This
enables modelling longer range dependencies within the model. These HOP
act as soft constraints during the optimisation. \citet{kohli2009} use a
$P^n$ Potts model-based CRF with HOP for the task of image segmentation
and use a graph cut based algorithm to solve the optimisation problem.
\citet{tarlow2010} proposed a method with HOP models and belief
propagation,
adopting a set of potentials for which efficient
message passing rules exist. In \citep{komodakis2009} a dual
decomposition based master-slave framework is presented to solve generic
higher order Markov random fields. While HOP can be created from the
same information as the constraints, they only form soft
constraints and as such can be violated in the final solution. Our
approach, in contrast, ensures that the imposed constraints from depth information
are satisfied by the solution. Further we exploit much simpler features compared to the state of the art methods while providing higher accuracy for even tricky classes such as pedestrians. Our method also has the potential to be implemented in real time.

\begin{figure}[h]
    \centering
    \includegraphics[width=0.8\linewidth]{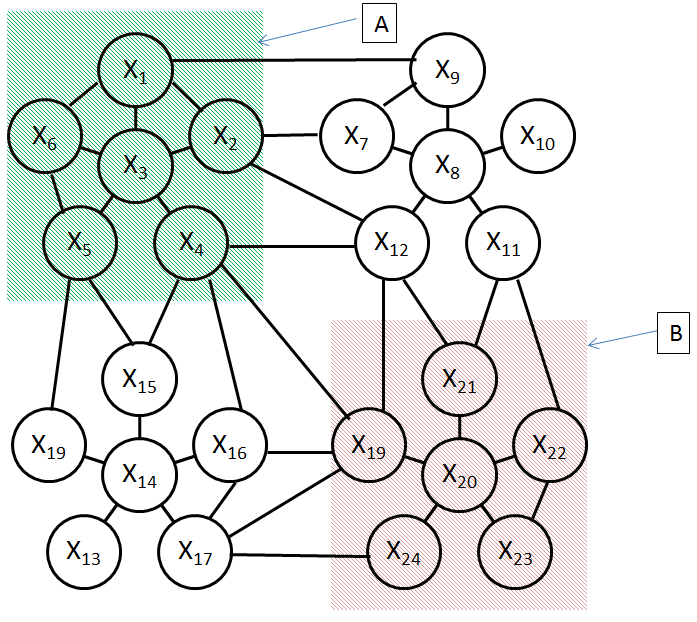}

    \caption{Example of the type of CRF graph used in this paper.
        Pairwise potentials are indicated by the edges, while the
        additional constraints are indicated by the two shaded areas, A
        and B. These areas encode sets of nodes which are required to
        be assigned the same label.}
    \label{fig:graph-example}
\end{figure}
\section{Globally Constrained CRF}
\label{sec:method}

Our segmentation method is based on conditional random fields (CRF) with
unary and pairwise potentials. The additional \emph{a priori}
information about sets of points which belong to the same group is
encoded as constraints on the CRF. A graphical representation of this
structure is shown in \figref{graph-example}, where nodes are denoted by
 circles while  edges indicate connections between nodes. The two
sets of nodes coloured identically represent sets of nodes constrained
to take the same label. The unary and pairwise potentials are based on
information extracted from the image while the information about groups
of nodes is extracted from 3D laser data. Our goal is to find the best
label assignment for each node, i.e.\ the MAP solution of the CRF. To do
this efficiently we represent the CRF as  a quadratic program.

\subsection{Conditional Random Field}

The log likelihood model of a conditional random field is given by:
\begin{multline}
    \log P \left( X \mid S \right) =
        \sum_{i \in S} \phi_i(X_i) \\
        + \sum_{i \in S, j \in \mathcal{N}(i)} \psi_{ij}(X_i, X_j) - Z(S),
    \label{eq:loglik}
\end{multline}
where $Z(S)$ is the normaliser, $X = \{X_1, X_2, \dots, X_N\}$ is the
set of discrete random variables associated with the super pixels \citep{sp} set $S$ in
the input image. Each super pixel $X_i$ is assigned one of the
output labels $L = \{1, \dots, K\}$. The potential functions of the CRF
are denoted by $\phi_i(X_i)$ for the unary potential and $\psi_{ij}(X_i,
X_j)$ for the pairwise potential defined for each super pixel $i$ and
each of its neighbours $\mathcal{N}(i)$.

\subsection{Quadratic Program Formulation}

The goal is to find the best assignment of labels to the nodes (MAP
assignment) considering local and global information. As finding the MAP
solution to \eqref{loglik} is NP hard we start by representing it as a
quadratic integer program of the following form:
\begin{subequations}
\label{eq:qp-integer}
\begin{align}
    \text{maximise}   & \quad
        \sum_{i\in S} \sum_{p \in L} \phi_i(x_i^p) \mu_i(x_i^p) \notag \\
                      & \quad + \sum_{\substack{i\in S\\ j\in \mathcal N(i)}} \sum_{p, q \in L} \psi_{ij}(x_i^p, x_j^q) \mu_i(x_i^p) \mu_j(x_j^q) \\
    \text{subject to} & \quad \sum_{p\in L} \mu_i(x_i^p) = 1 \quad \forall i \label{eq:qp-2} \\
                      & \quad \mu_i(x_i^p) \in \{0, 1\} \quad \forall i, p \label{eq:qp-3},
\end{align}
\end{subequations}
with the indicator function:
\begin{equation}
    \mu_i(x_i^p) = \begin{cases}
        1 & \text{if } (X_i = p ) \wedge (x_i^p = 1) \\
        0 & \text{otherwise}
    \end{cases},
    \label{eq:indicator}
\end{equation}
where $x_i^p$  encodes if node $X_i$ has been assigned label $p$. This
quadratic program formulation penalises disagreements between the data
via the indicator function, which guides the model to obtain coherent
segmentations. Additionally, \eqref{qp-2, qp-3} enforce that exactly one
label is selected for each node. Relaxing the integer requirement of the
quadratic program \citep{Zhang2012} we obtain:
\begin{subequations}
\label{eq:qp-relax}
\begin{align}
    \text{maximise}   & \quad
        \sum_{i\in S} \sum_{p\in L} \phi_i(x_i^p) \mu_i(x_i^p) \notag \\
                      & \quad + \sum_{\substack{i\in S\\ j\in N(i)}} \sum_{p, q\in L} \psi_{ij}(x_i^p, x_j^q) \mu_i(x_i^p) \mu_j(x_j^q) \\
    \text{subject to} & \quad \sum_{p\in L} \mu_i(x_i^p) = 1 \quad \forall i \\
                      & \quad 0 \leq \mu_i(x_i^p) \leq 1 \quad \forall i, p.
\end{align}
\end{subequations}

Optimising \eqref{qp-relax} yields an approximation to the MAP
solution for the segmentation problem. However, it does not yet include
the global constrains on sets of nodes. Adding these constraints we
obtain:
\begin{subequations}
\label{eq:qp-relax-global}
\begin{align}
    \text{maximise}   & \quad
        \sum_{i\in S} \sum_{p\in L} \phi_i(x_i^p) \mu_i(x_i^p) \notag \\
                      & \quad + \sum_{\substack{i\in S\\j\in N(i)}} \sum_{p, q\in L} \psi_{ij}(x_i^p, x_j^q) \mu_i(x_i^p) \mu_j(x_j^q)  \label{cof}  \\
    \text{subject to} & \quad \sum_{p\in L} \mu_i(x_i^p) = 1 \quad \forall i \\
                      & \quad \sum_{i, j \in C_k} \sum_{p \in L} \mu_i(x_i^p) - \mu_j(x_j^p) = 0
                        \quad \forall C_k \in \mathcal C \label{eq:qp-g-2} \\
                      & \quad 0 \leq \mu_i(x_i^p) \leq 1 \quad \forall i, p
\end{align}
\end{subequations}
where \eqref{qp-g-2} enforces that all pairs of points $i$ and $j$ in
a constraint set $C_k \in \mathcal C$ are assigned the same label.

In order to solve \eqref{qp-relax-global} efficiently we follow
\citep{Krogstad2012} and rewrite it in matrix notation:
\begin{subequations}
\label{eq:matrix-representation}
\begin{align}
    \text{maximise}     & \quad \frac{1}{2} A^T Q A + b^T A \\
    \text{subject to}   & \quad EA = d   \label{equal} \\
                        & \quad 0 \leq A \leq 1,
\end{align}
\end{subequations}
where $Q$ encodes the quadratic coefficients (pairwise potentials) and
$b$ the linear coefficients (unary potentials). $A$ is the indicator
matrix representing the $\mu_i(x_i^p)$ variables and $E$ encodes the
global constraints from \eqref{qp-g-2}. The solution to
\eqref{matrix-representation} can be found by introducing Lagrange
multipliers as follows:
\begin{subequations}
\begin{align}
    \text{maximise} & \quad \frac{1}{2} A^T Q A + b^T A + \lambda E A\\
    \text{subject to}      & \quad 0 \leq A \leq 1,
\end{align}
\label{eq:lagran}
\end{subequations}

We can achieve the same maximum as in \eqref{matrix-representation} by
making $\lambda E A$ equal to zero. To this end we introduce new
variables:
\begin{align}
    E Z = 0 \label{eq:nulsp} \\
    Z R = A \label{eq:newy},
\end{align}
where $R$ has the dimension $dim(A) - dim(E)$, while solving
\eqref{nulsp} implies that $Z$ is the null space of $E$. Substituting
these two equations back into \eqref{lagran} we obtain:
\begin{subequations}
\begin{align}
    \text{maximise}   & \quad \frac{1}{2} R^T (Z^T Q Z) R + (Z^T c)^T R \\
    \text{subject to} & \quad 0 \leq R \leq 1
\end{align}
\label{eq:uncons}
\end{subequations}
This transformation has two benefits: First, the dimensionality of $R$ is
reduced compared to that of $A$ based on the number of constraints. This
means that a large number of constraints makes the optimisation problem
easier to solve. Second, the optimisation problem is now unconstrained
which again makes it easier to solve.

Similar to the transformation from \eqref{qp-relax-global} to
\eqref{matrix-representation} we can rewrite
\eqref{uncons} using element wise notation as follows:
\begin{subequations}
\label{eq:lagrange-dual-optimisation}
\begin{align}
    \text{maximise}   & \quad
        \sum_i \sum_p \rho_i(y_i^p) \mu_i(y_i^p) \notag \\
                      & \quad + \sum_{i, j} \sum_{p, q} \tau_{ij}(y_i^p, y_j^q)
                            \mu_i(y_i^p) \mu_j(y_j^q) \label{eq:qp-lagrange-1} \\
    \text{subject to} & \quad 0 \leq \mu_i(y_i^p) \leq 1,
\end{align}
\end{subequations}
with the unary potential $\rho_i = - Z^T c$ and the pairwise potential
$\tau_{ij} = - Z^T Q Z$ and  $y_i^p$ denotes
if label $p$ has been assigned to node $Y_i$. We optimise
\eqref{lagrange-dual-optimisation} using gradient ascent which can be
done efficiently as the gradient can be computed in closed form
\citep{Zhang2012}:
\begin{align}
    \begin{split}
    q_i(y_i^p) & = \frac{\partial B}{\partial \mu_i(y_i^p)} \\
               & = \rho_i(y_i^p) + 2 \sum_{i, j} \sum_{q}
                        \tau_j(y_i^p, y_j^q) \mu_j(y_j^q)
    \label{eq:san}
    \end{split} \\
    \mu_i^{t+1}(y_i^p) & =
        \frac{\mu_i^t(y_i^p) q_i(y_i^p)}{\sum_q \mu_i^t(y_i^q) q_i(y_i^q)},
    \label{eq:man}
\end{align}
with $B$ standing for \eqref{qp-lagrange-1}.

This allows us to implement a highly efficient gradient ascent based
algorithm as the gradient can be evaluated directly in closed form. Once
the algorithm has converged we can extract the values of
original  indicator variables $\mu(x_i^q)$ and thus the MAP label assignments to the $X_i$
variables. To this end we transform the solution for $\mu(y_i^p)$ obtained
from \eqref{lagrange-dual-optimisation} back into the form of
\eqref{qp-relax-global} using $A = Z R$. $A$ is a column vector whose
entries correspond to the values of the $\mu(x_i^p)$. The optimal assignment
to each node $X_i$ is found by selecting the label $p \in L$ for
which $\mu(x_i^p) = 1$ holds. This is summarised in \algoref{gcrf-algo}.
The required inputs are the values of the potentials over the possible
$R$ value settings. Then the gradient(\eqref{san})is computed and used to
update the solution iteratively until convergence is achieved. Finally,
the solution is extracted and returned.\\

\begin{algorithm}
    \caption{Globally Constrained CRF}
    \label{algo:gcrf-algo}

    \KwIn{Potential values $\rho$ and $\tau$}
    \KwOut{Assignment of $X$}

    \tcp{Perform gradient descent}
    \Repeat{convergence}
    {%
        \ForEach{$i \in \{1, \dots, N\}$}
        {%
            \ForEach{$p \in \{1, \dots, L\}$}
            {%
                $q_i(y_i^p) \leftarrow$
                    \\\quad
                    $\rho_i(y_i^p) + 2 \sum_{i, j} \sum_{q}
                    \tau_j(y_i^p, y_j^q) \mu_j^t(y_j^q)$
                $\mu_i^{t+1}(y_i) \leftarrow
                    \frac{\mu_i^t(y_i) q_i(y_i)}{\sum_i \mu_i^t(y_i) q_i(y_i)}$
            }
        }
    }

    \tcp{Extract final solution}
    $A \leftarrow Z R$\\
    \ForEach{$i \in \{1, \dots, N\}$}
    {%
        \ForEach{$p \in \{1, \dots, L\}$}
        {%
            $X_i \leftarrow p \text{ if }\mu( x_i^p )= 1$
        }
    }

    \KwRet{$X$}
\end{algorithm}
\begin{table}[bt]
\footnotesize
    \centering

        \begin{tabular}{llr}
        \toprule
        Type & Description & Dimensionality \\
        \midrule
        Texture              & RGB gradient magnitude histogram   & $50 \times 3 = 150$ \\
                             & RGB gradient orientation histogram & $50 \times 3 = 150$ \\
        Colour               & RGB mean                           & 3 \\
                             & RGB std                            & 3 \\
                             & HSV histogram                      & $50 \times 3 = 150$ \\
        Location             & Super pixel image coordinates      & $200$ \\
        \bottomrule
    \end{tabular}
  \caption{Features used for the unary potential of the CRF based on a
        discriminant analysis classifier applied to super pixels.}
    \label{tab:features}
\end{table}
\section{Experiments}
\label{sec:experiments}

In this section we present experimental evaluation of our proposed
framework on the task of image-based scene segmentation. We use the
KITTI dataset \citep{Geiger2011} as it provides typical urban data.
The dataset was captured by driving around the city of Karlsruhe.
Importantly, the data contains both colour images and Velodyne depth
data. The image information is used to build the CRF model structure and
potential functions while the Velodyne data is used to construct
global constraint sets.

\subsection{Model Building}
\label{sec:dataset}

\begin{figure}
    \centering
     \vspace{3mm}
      \subfloat{
        \includegraphics[width=0.85\linewidth]{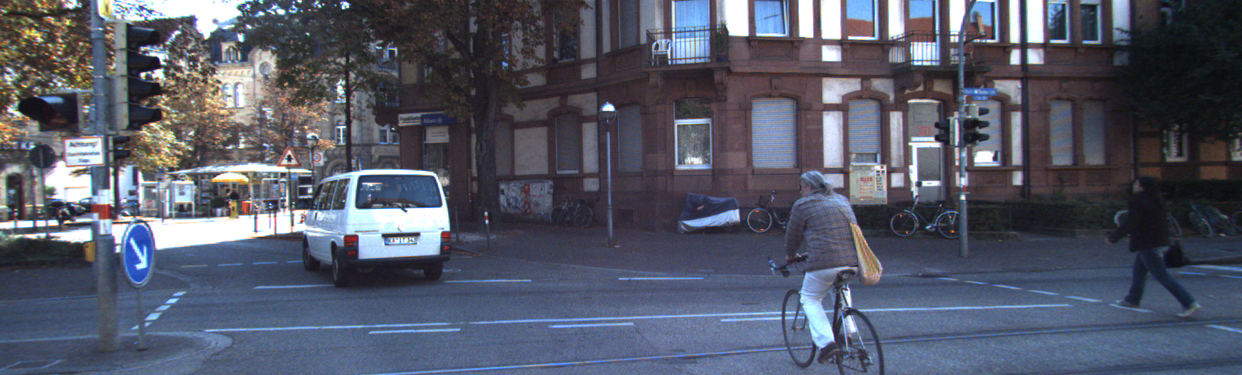}
        \label{fig:example-orig}
    } \\
    \vspace{-2mm}
        \subfloat{
        \includegraphics[width=0.85\linewidth]{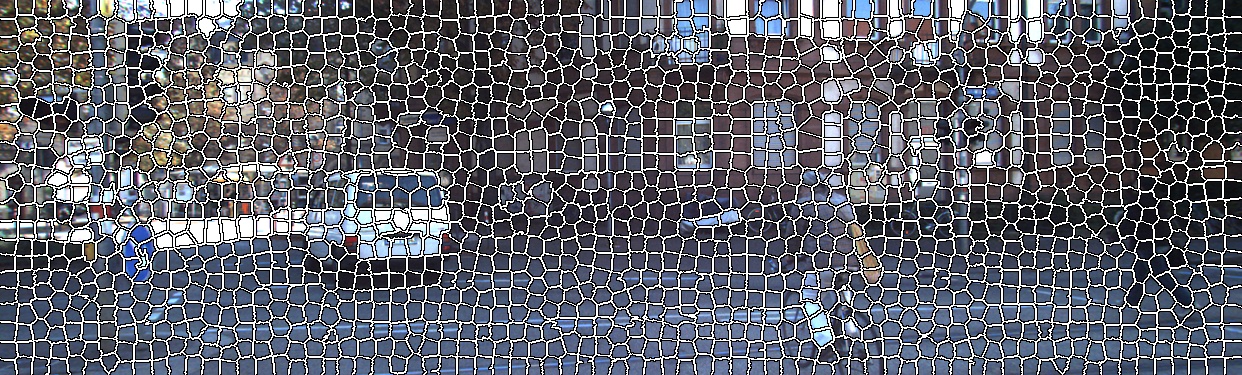}
        \label{fig:example-sp}
    } \\
    \vspace{-2mm}
    \subfloat{
        \includegraphics[width=0.85\linewidth]{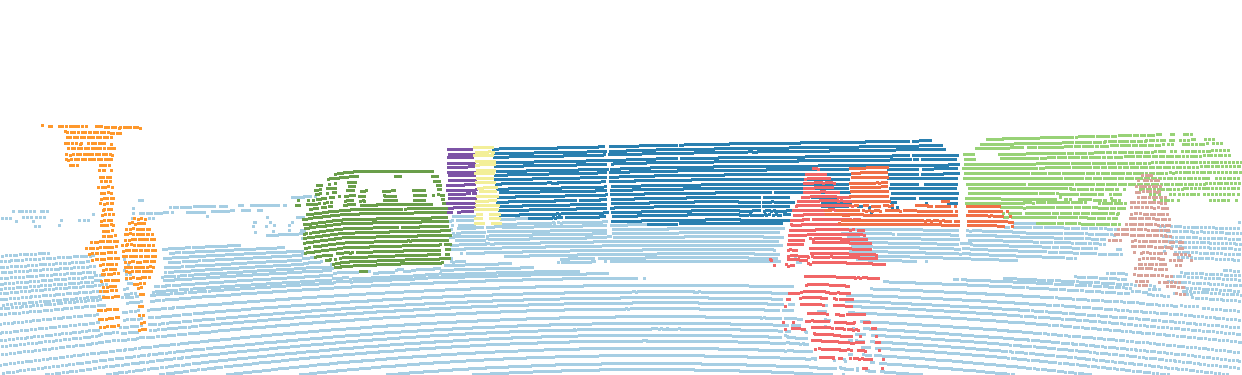}
        \label{fig:example-constraints}
    }
    \caption{Display of a typical scene from the KITTI dataset.top image shows
        the raw image while middle overlays the super pixels extracted from
        the image. In bottom image the global constraints extracted from the 3D
        laser point clouds are shown projected into the image space,
        each colour represents a single segment. Labels of the segments
        are unknown at this stage.}
\end{figure}

We start by extracting super pixels from the image using SLIC
\citep{Achanta2010} which forms an over segmentation of the original
image. From each $375 \times 1242$ image we extract roughly \num{1600}
super pixels, shown in \figref{example-sp}. Each of these represents a
node in our CRF and the goal is to label them with one of the seven
different classes: vehicle, pedestrian \& cyclist, buildings, road \&
paved area, sky, vegetation, and unknown. Due to the low sample size of
pedestrians and cyclists in the dataset they are assigned to same
class. The edges between nodes are defined by their distance within the
image, i.e.:

\begin{equation}
    E(i, j) = \begin{cases}
        1 & \text{if } \text{dist}(i, j) < \Theta \\
        0 & \text{otherwise}
    \end{cases},
\end{equation}
where $\Theta$ is the distance threshold and $\text{dist}(i, j)$ is the
Euclidean distance in image coordinates between centres of two super
pixels. All super pixels closer than the user defined distance $\Theta$ are
connected.  In our experiments $\Theta$ was set such that each node is
connected to roughly ten neighbouring nodes, which results in a roughly
grid like structure. 

The unary potentials $\phi_i$ are obtained from the posterior of a
pseudo linear discriminant analysis classifier \citep{mika1999}. The
classifier is trained on $120$ manually labelled images from the KITTI
dataset using colour, texture, and location features, shown in
\tabref{features}.
The pairwise potentials $\psi_{ij}$ are derived based on their
dissimilarity using colour, texture, and location information of the super
pixels, i.e.:
\begin{equation}
\begin{split}
\text{dis}(i, j) = &
            D(\text{colour\_hist}(i), \text{colour\_hist}(j) \\
        & + \theta_c ||\text{mean\_colour}(i) - \text{mean\_colour}(j)||_2 \\
        & + \theta_l ||\text{com}(i) - \text{com}(j)||_2
        ) / 3,
\end{split}
\end{equation}
where $\text{mean\_colour}(i)$ is the mean colour of the $i$-th super
pixel normalised to $1$ by $\theta_c$, $\text{com}(i)$ is the centre of
mass of the super pixel in pixel coordinates, normalised to $1$ with
$\theta_l$, and $\text{colour\_hist}(i)$ is the colour histogram of the
$i$-th super pixel whose difference is computed using the Bhattacharya
distance: 
\begin{equation}
\footnotesize
    D(a, b) = \sqrt{%
        1 - \frac{1}{\sqrt{\sum_i a_i \sum_i b_i N^2}}
        \sum_i \sqrt{a_i b_i}
    },
    \label{eq:bhattacharyya}
\end{equation}
where $a$ and $b$ are two histograms and $N$ is the number of bins in
the histograms. This results in a similarity value between \num{0} and
\num{1}, with $0$ encoding identical super pixels. As the constraint
function requires a value of $1$ for identical super pixels we use the
following final pairwise potential function:

\begin{equation}
\footnotesize
    \psi_{i, j}(x_i^p, x_j^q) = \begin{cases}
        1 - \text{dis}(i, j)^2 & \text{if } p = q \\
        \text{dis}(i, j)^2     & \text{otherwise}
    \end{cases}.
\end{equation}

We obtain the global constraints on sets of super pixels by extracting
groups of connected points, or objects, from the Velodyne point cloud.
This is facilitated by the KITTI dataset providing time synchronised
camera images and Velodyne point clouds. To this end we first perform a
simple ground plane removal step using RANSAC to find the largest plane
aligned with the ground. The remaining points are then grouped using
Euclidean distance based clustering \citep{pcl}. This results in a
collection of clusters, of which we only consider those that contain
more then \num{150} points which ensures that each clusters contains
only points belonging to a single class. The ground plane as well as the
retained segments of this process can be seen in
\figref{example-constraints}. The 3D coordinates of the points contained
in the selected clusters are then translated into image space
coordinates using the extrinsic calibration provided by the KITTI
dataset and then associated with super pixels. Based on this mapping we
create the constraint sets $\mathcal C$ used in the optimisation. All
super pixels that correspond to the same laser segment are constrained
to be assigned the same label. Super pixels which do not belong to any
of the extracted laser segments are kept unconstrained.

\subsection{Segmentation Quality}

In the following we present image only CRF solutions obtained using
loopy belief propagation (LBP) and quadratic programming (QP) to
showcase the quality of the results obtained by these methods without
using any additional constraints. Thereafter, we introduce the
constraints obtained from the Velodyne and compare the results obtained
using a graph-cut based HOP method \citep{kohli2008} with our hard
constraint based method.

\noindent\emph{\textbf{Visual Information Only Segmentation}}

\begin{figure*}[bt]
    \centering
    \includegraphics[width=0.8\textwidth]{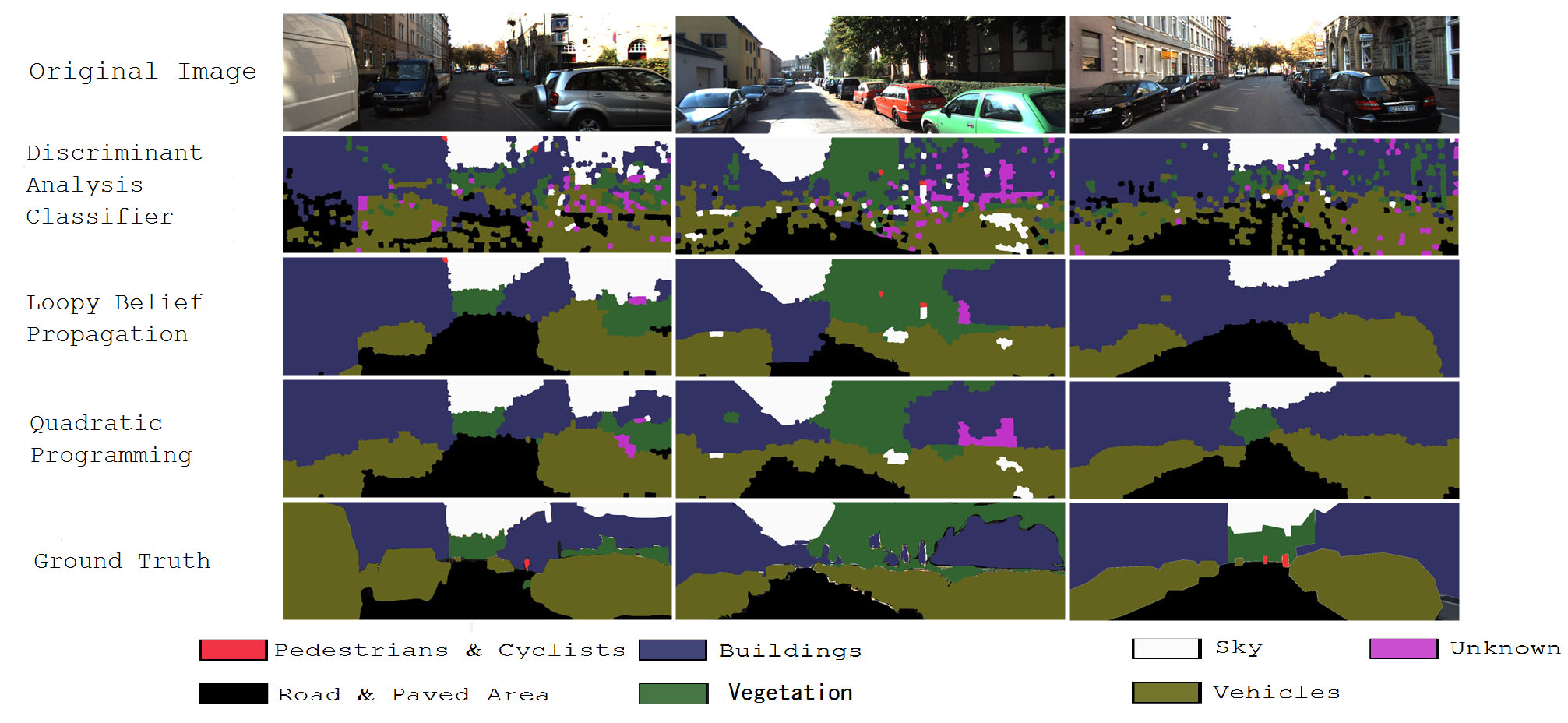}

    \caption{This figure shows exemplary scene segmentation results
        obtained on several images. From top to bottom we have: the
        original image processed by each method, discriminant analysis
        classifier, loopy belief propagation, quadratic programming, and
        ground truth.}
    \label{fig:visual-only-results}
\end{figure*}

We present results from three methods, (i) discriminant analysis
classifier which provides the unary potentials of the CRF, (ii) loopy
belief propagation using the UGM toolbox \citep{ugm}, and (iii)
quadratic programming solution \citep{Zhang2012}. Exemplary results
together with the original image and ground truth labels are shown in
\figref{visual-only-results}. The first row shows the original colour
images while the second row shows the most likely class of the
discriminant analysis classifier which is used as the unary potentials
of the CRF. As to be expected the classifier output is noisy and
incorrect in several places. Both the LBP and QP based CRF solutions
produce a much cleaner and consistent result compared with the raw
classifier result. However, there are still segmentation errors present
due to effects such as shadowing and illumination changes. The
quantitative evaluation results from \num{100} manually labelled images,
shown in \tabref{method_evaluation}, further demonstrates the
improvements and also indicates that the QP based solution outperforms the
LBP one. This demonstrates that the basis on which our method is built
is capable of producing high quality segmentation results before any
additional constraints are added, which will be evaluated next.

\noindent\emph{\textbf{Laser Constrained Segmentation}}

In this section we explore the impact additional constraints, extracted
from Velodyne data, have on segmentation results by comparing our method
to a HOP based method by \citet{kohli2008}. The higher order potentials
penalise label inconsistencies between nodes identified to be part of a
single segment in the 3D data. Both methods use uniform weight
parameters for the unary, pairwise, and higher order potentials, where
applicable.

\begin{figure*}[bt]
    \centering

    \includegraphics[width=1\textwidth]{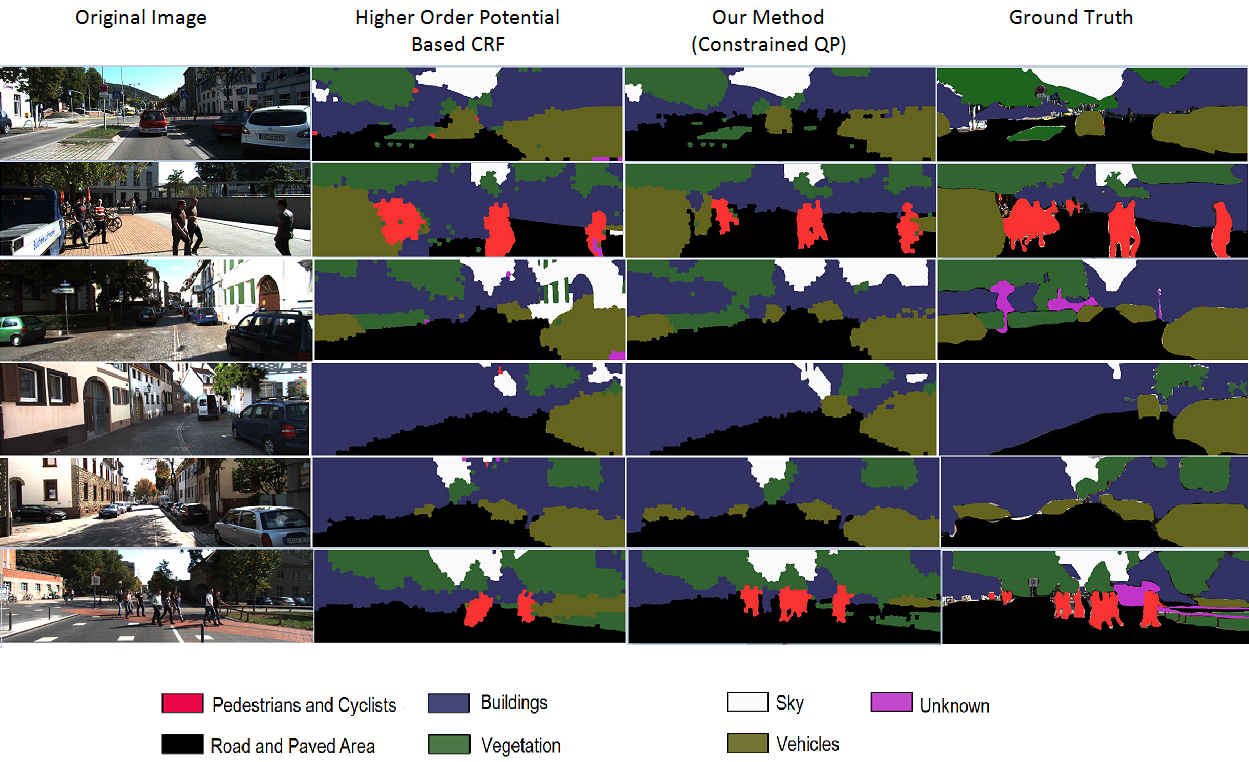}

    \caption{This figure shows results obtained on various scenes using
        the HOP based method and our proposed method together with the
        original image and the hand labelled ground truth. We can see that
        our method (Constrained QP) performs better then the HOP based
        method at segmenting distant and objects cast in shadows.}
    \label{fig:constraints-based-results}
\end{figure*}

\begin{figure*}[bt]
    \centering
  \vspace{2mm}
    \includegraphics[width=0.9\textwidth]{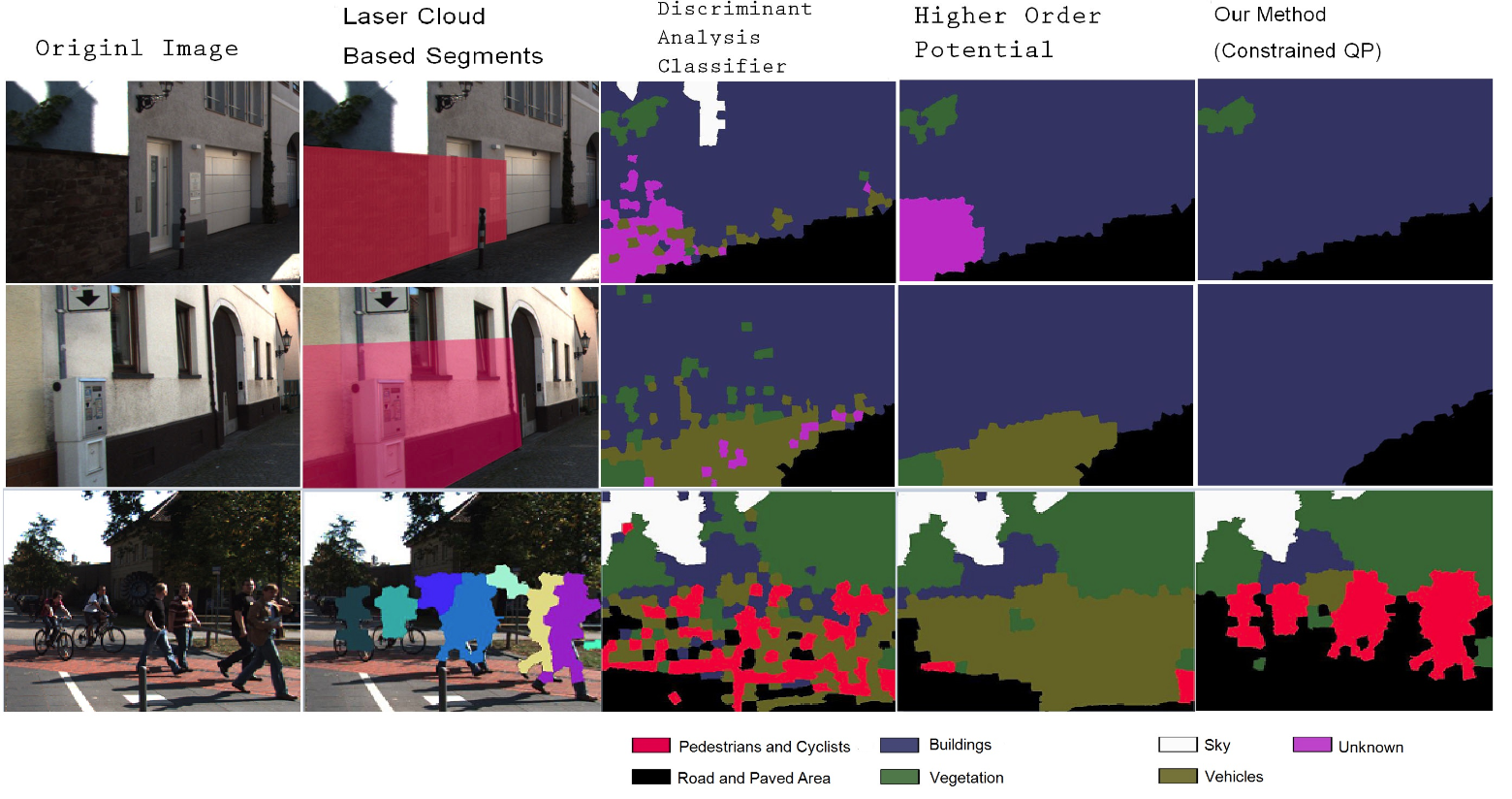}

    \caption{Examples of the benefits that enforcing hard constraints
        provide. The highlighted areas in the image show continuous 3D
        segments extracted from Velodyne data. The classifier output in
        these areas is noisy and wrong due to visual ambiguities. While
        the HOP based method fails to correct this our method succeeds
        in classifying the entire area correctly, as it is forced to
        assign a single class to each of the laser based segments.}
    \label{fig:constraint-differences}
\end{figure*}

Some exemplary results are shown in \figref{constraints-based-results}
with the original image shown on the far left, followed by the result of
the HOP based method in the second column, then our method, and finally
the hand labelled ground truth. Inspecting the results we can see that
the HOP based method struggles to correctly identify distant objects,
especially when cars or walls are involved. Additionally, the results
our method obtains appear more uniform with less spurious
classifications. This difference in behaviour is explained by the way
the additional 3D information is used. While our method enforces the
constraints the HOP based method is allowed to violate them. The
examples in \figref{constraint-differences} show the benefit of using
the hard constraints rather then soft constraints. The first two rows
showcase this for a single wall while the third row shows the result of
this in a scene populated by pedestrians. The first two columns show the
original image and the segment extracted from the Velodyne data. Due to
the visual appearance of these areas the classifier fails to pick the
correct class in some parts of the 3D segment. The HOP based method
fixes some classification errors, however, cannot fix every single one.
In the case of the pedestrian scene the HOP method even misclassifies
all pedestrians. Our method on the other hand is forced to assign a
single class to the entire segment and as such the correct class is
assigned even to the areas where the classifier makes mistakes.

For a quantitative analysis we compute average precision, recall,
accuracy, and F1-score for the different methods on \num{100} labelled
images. As we can see in \tabref{method_evaluation} the addition of
global constraints in our method allows it to significantly outperform
the other methods lacking this information and even the HOP method,
using the same information, does not provide the same benefits. This
shows that adding constraints based on simple information about which
areas belong to a single object allows the segmentation to be more
accurate. This is good news, as this type of information is readily
available in robotic systems. Looking at the performance of the
individual classes in \tabref{class_evaluation} we can see that
``cyclist \& pedestrian'' class is the hardest one. This is explained by
the fact that instances of this class occur infrequently and as such the
classifier has a harder time at classifying them correctly. Furthermore,
this class has the smallest appearance in the Velodyne data and as such
will only be detected at close range. The other classes exhibit similar
performance, which is not surprising, given that they occur frequently
in the data and cover larger areas of the scene.

The performance of both constrained QP and HOP can be improved by
training the weight parameters of the potential functions, which encodes
knowledge about class relationships and object co-occurrence statistics.
The advantage of our method is, that it only requires unary
and pairwise potentials while HOP has additional higher order
potentials, which can be harder and time consuming to learn. This makes
the proposed method easier to fine tune as there are fewer parameters
involved.

\begin{table*}[bt]
    \centering

    \begin{tabular}{lrrrr}
        \toprule
        Method & Average Precision & Average Recall & Average Accuracy & F1 Score\\
        \midrule
        Discriminant Analysis Classifier          & $0.7027 \pm 0.045$ & $0.5127 \pm 0.061$ & $0.8826 \pm 0.045$ & $0.5927 \pm 0.057$ \\
        Loopy Belief Propagation                  & $0.7435 \pm 0.051$ & $0.7197 \pm 0.081$ & $0.9024 \pm 0.053$ & $0.7314 \pm 0.067$ \\
        Quadratic Programming Relaxation          & $0.8001 \pm 0.032$ & $0.7645 \pm 0.048$ & $0.9150 \pm 0.053$ & $0.7818 \pm 0.040$ \\
        \midrule
        Higher Order Potentials \citep{kohli2008} & $0.8319 \pm 0.073$ & $0.8143 \pm 0.067$ & $0.9278 \pm 0.022$ & $0.8230 \pm 0.070$ \\
        Constrained Quadratic Programming         & $0.8549 \pm 0.079$ & $0.8424 \pm 0.078$ & $0.9507 \pm 0.025$ & $0.8482 \pm 0.076$ \\
        \bottomrule
    \end{tabular}

    \caption{Quantitative evaluation of various segmentation methods.
        The first three rows represent standard CRFs using only image
        based information. The last two rows show the results for
        methods using additional information obtained from 3D Velodyne
        scans. The HOP method incorporates this information as an
        additional potential, while our method (Constrained Quadratic
        Programming) enforces the validity of this additional
        information as constraints. We can see that the addition of the
        3D information improves the performance compared to the image
        only based solutions. However, actively enforcing the
        constraints allows our method to outperform the HOP based
        method.}
    \label{tab:method_evaluation}
\end{table*}

\begin{table*}[bt]
    \centering

    \begin{tabular}{rrrrrrrrr}
        \toprule
        \multicolumn{1}{c}{Quality Measure} & \multicolumn{2}{c}{Average Precision} & \multicolumn{2}{c}{Average Recall} & \multicolumn{2}{c}{Average Accuracy} & \multicolumn{2}{c}{F1 Score} \\
        \midrule
        \multicolumn{1}{c}{Method} & {HOP}& {CQP} & {HOP}& {CQP} & {HOP} & {CQP }& {HOP}& {CQP} \\
        \midrule
        Cyclists \&Pedestrians    & $0.7689$ & $0.7700$ & $0.5134$ & $0.5334$ & $0.9670$ & $0.9772$ & $0.6153$ & $0.6302$ \\
        Roads \& Paved Area       & $0.8431$ & $0.8554$ & $0.9747$ & $0.9775$ & $0.9569$ & $0.9789$ & $0.9032$ & $0.9124$ \\
        Vegetation                & $0.8284$ & $0.8440$ & $0.5161$ & $0.5359$ & $0.9468$ & $0.9473$ & $0.5931$ & $0.6192$ \\
        Buildings                 & $0.8431$ & $0.8420$ & $0.8448$ & $0.8838$ & $0.8652$ & $0.9103$ & $0.8382$ & $0.8568$ \\
        Sky                       & $0.7519$ & $0.7877$ & $0.7690$ & $0.7564$ & $0.9723$ & $0.9780$ & $0.7265$ & $0.7461$ \\
        Vehicles                  & $0.8485$ & $0.9089$ & $0.7101$ & $0.8058$ & $0.9031$ & $0.9413$ & $0.7614$ & $0.8543$ \\
        \bottomrule
    \end{tabular}

    \caption{Quantitative evaluation of the performance on a per class
        for HOP method and our method CQP. All the major 6 classes
        excluding the unknown class are separately evaluated for the
        quality of segmentation.}
    \label{tab:class_evaluation}
\end{table*}

\subsection{Runtime Comparison}

\begin{figure}
    \centering
    \vspace{2mm}
    \input{speed_comparison.tex}
    \caption{The plot shows the time (log scale) needed to find a
        solution as a function of the number of nodes in the CRF. NLopt
        BOBYQA solving the problem directly scales very poorly while our
        proposed method is scaling much more favourably.}
    \label{fig:runtime}
\end{figure}
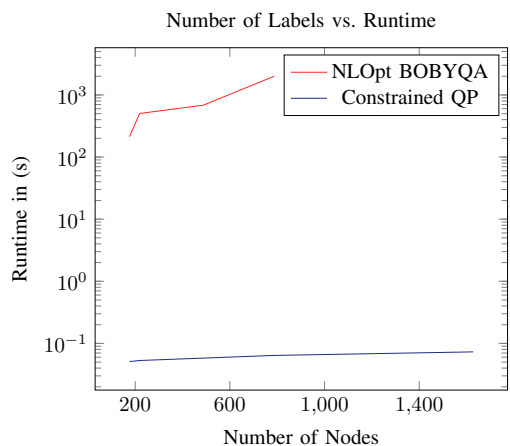

We start by comparing the runtime required to solve the constrained quadratic
program of \eqref{qp-relax-global} directly using NLOPT
BOBYQA \citep{nlopt} compared to our proposed framework.
As we can see in \figref{runtime}, directly solving the
quadratic program is not feasible for problems of interesting size. On
the other hand, our method scales very favourably with the problem size.
Additionally, while typically increasing the number of constraints makes
the problem harder and thus slower to solve, our method becomes faster
with more constraints. This is caused by the fact that constraints
reduce the size of the actual problem we solve. This means that adding
more domain knowledge allows us to improve the quality of the result as
well as speed up the computation.

A typical CRF derived from the images used in the experiments consists
of \num{1600} nodes, each of which can have one of seven different
labels, which means we have on the order of \num{11200} random
variables. Solving this CRF using the quadratic program formulation \eqref{qp-relax}
(with no laser based constraints) 
takes around \SI{2}{\second} while the belief propagation based
solution takes \SI{0.5}{\second}. Including the constraints we can
reduce the number of nodes to around \num{400} which results in a much
smaller number of variables, around \num{2800}. Solving this problem
using  gradient based method takes around \SI{0.07}{\second}. All
computations were performed on an Intel Core i5 3.20 GHz processor with
C++ implementations of the algorithms. Besides the reduction of the
number of variables involved our method also requires fewer iterations
to converge, around \num{25}, compared to \num{70} for the purely image
based quadratic program. These two advantages, reduction in number of
variables and faster convergence gives our method a significant
computational advantage.

\section{Conclusion}
\label{sec:conclusion}

In this paper we presented a novel image segmentation method based on a
conditional random field with additional global constraints which encode
\emph{a priori} information about groups of nodes having the same label
obtained from a secondary sensor. This CRF is formulated as a
relaxed quadratic program whose MAP solution is found using gradient
descent based optimisation. We evaluate our method on data from the
KITTI project. Each image is pre-processed into super pixels which
provide the unary and pairwise potentials of the CRF. The global
constraints on sets of super pixels are obtained from Velodyne data. The
results show that the addition of these hard constraints significantly
improves on the solution obtained without constraints. Runtime
comparisons show how black box solvers do not scale for this problem and
how our formulation exploits constraints in a way which simplifies the
problem. Finally, the proposed method is general and capable of encoding
other forms of constraints, such as relative positioning of classes with
respect to each other.

\vspace{-3mm}
\renewcommand{\bibname}{\section{References}}
\renewcommand{\bibfont}{\small}
{\bibliographystyle{plainnat}}
\bibliography{forbib}

\end{document}

%% file: speed_comparison.tex
\begin{tikzpicture}[scale=0.8]

    \begin{axis}[
        xlabel=Number of Nodes,
        ylabel=Runtime in (\si{\second}),
        title=Number of Labels vs. Runtime,
        xtick={200, 600, 1000, 1400, 1800},
        ymode=log
    ]

        \addplot[col1] plot coordinates {
            (176, 213)
            (219, 503)
            (488, 684)
            (787, 2000)
        };
        \addplot[col2] plot coordinates {
            (176, 0.051)
            (219, 0.053)
            (787, 0.064)
            (1628, 0.073)
        };

        \legend{NLOpt BOBYQA, Constrained QP};

    \end{axis}
\end{tikzpicture}